\documentclass[letterpaper, 10 pt, conference]{ieeeconf}  

\IEEEoverridecommandlockouts                              

\usepackage{graphics} 
\usepackage{epsfig} 
\usepackage{amsmath} 
\usepackage{amssymb}  
\usepackage{url}

\usepackage{flushend}
\usepackage{float}
\usepackage{xcolor}

\title{\LARGE \bf
Probabilistic Contact State Estimation for Legged Robots using Inertial Information}

\author{ Michael Maravgakis$^{1}$, Despina-Ekaterini Argiropoulos$^{1}$, Stylianos Piperakis$^{2}$ and Panos Trahanias$^{1}$
\thanks{*The research leading to these results has received partial funding from the European Community’s HORIZON.1.2 - Marie Skłodowska-Curie Actions (MSCA) under Grant agreement No. 101072634, project RAICAM and the HYPERION project, funded from the Prefecture of Crete under KPHP1-0033203, act code (MIS) ``5063282''.}
\thanks{$^{1}$ Michael Maravgakis, Despina-Ekaterini Argiropoulos and Panos Trahanias are with the Institute of Computer Science, Foundation
for Research and Technology - Hellas (FORTH), Heraklion, Greece. {\tt\small \{maravgakis,despinar,trahania\}@ics.forth.gr} }
  \thanks{$^{2}$ Stylianos Piperakis is with the Ownage Dynamics L.P. {\tt \small stpiperakis@gmail.com}}
}%

\begin{document}

\maketitle
\thispagestyle{empty}
\pagestyle{empty}

\begin{abstract}
Legged robot navigation in unstructured and slippery terrains depends heavily on the ability to accurately identify the quality of contact between the robot's feet and the ground. 
Contact state estimation is regarded as a challenging problem and is typically addressed by exploiting force measurements, joint encoders and/or robot kinematics and dynamics. 
In contrast to most state of the art approaches, the current work introduces a novel probabilistic method for estimating the contact state based solely on proprioceptive sensing, as it is readily available by Inertial Measurement Units (IMUs) mounted on the robot's end effectors. 
Capitalizing on the uncertainty of IMU measurements, our method estimates the probability of stable contact.
This is accomplished by approximating the multimodal probability density function over a batch of data points for each axis of the IMU with Kernel Density Estimation. 
The proposed method has been extensively assessed against both real and simulated scenarios on bipedal and quadrupedal robotic platforms such as ATLAS, TALOS and Unitree's GO1.

\end{abstract}

\section{Introduction}

The main advantage of legged robots against other mobile robots is their ability to discretize space~\cite{overviewLegged}, which enables them to traverse rough terrains, climb stairs and navigate in cluttered environments. 
To accomplish these tasks, a series of contacts with the ground needs to be planned and executed successfully~\cite{gaitPlanning1,gaitPlanning2}. 
Therefore, achieving agile locomotion and robust navigation requires accurate real-time knowledge of each foot's contact state (i.e. how firmly the foot is mounted on the ground).

The contact state depends directly on the Ground Reaction Force (GRF) and the friction coefficient between the foot and the ground according to the Coulomb's model for dry friction. 
The former is typically acquired directly from a Force Torque (F/T) sensor while the latter needs to be experimentally measured either by a human or a dedicated sensor. 
Often these sensors are impractical due to various reasons. 
For example, reliable force sensors are expensive and tend to degrade over repetitive use, as consequence of high speed motions and large impact wrenches. 
For the particular case of quadruped robots a further limitation regards the sensor's mass.
An average F/T sensor weights 100 g, thus accounting for approximately 50\% of the overall leg inertia, resulting to reduced acceleration capability~\cite{ProbabilisticFoot}.

Most approaches that deal with the contact detection problem can be classified into one of the following, model based or learning based approaches.
In general, learning based approaches require ground truth labels of the contact state which is often difficult to obtain. 
On the contrary, model based approaches do not suffer from this restriction. However, their main limitation lies in the modelling of the variables, which are assumed to follow predefined distributions.
By far the simplest and most widely used approach is to just threshold empirically the vertical GRF, which generally works when it is assumed that there is enough friction to prevent slippage. 
However, this heuristic is deemed to fail in cases where a foot slips and the vertical GRF is larger than the employed threshold. The latter inevitably leads to catastrophic results, and thus calls for more sophisticated approaches for contact state estimation.

Bloesch \textit{et al.}~\cite{bloesch1} proposed a state estimation framework that fuses inertial measurements with leg kinematics by employing an Extended Kalman Filter (EKF). 
The absolute position of each foothold is included in the state vector and the contact classification occurs by determining whether the pose of the foot is constant. 
Capitalizing on this work, the same group extended their research~\cite{bloesch2} by substituting the EKF with an Unscented Kalman Filter and reformulating the state vector to be expressed in the base frame. 
In order to robustify the filter, they also introduced an outlier rejection method in the update step.
They evaluated their approach on a StarlETH quadruped robot, proving its robustness to a certain amount of foot slippage detection. 
Hwangbo \textit{et al.}~\cite{ProbabilisticFoot} estimated the GRFs by exploiting kinematic and dynamic models. 
The binary contact state is subsequently exported from a Hidden Markov Model without any force sensing dependence. In~\cite{Kuindersma,Fallon,PiperakisRAL}, a Schmitt trigger (double threshold with hysteresis)  is employed to classify the contact of a humanoid robot. 
Although Schmitt trigger is a more sophisticated approach than just thresholding the vertical GRF, it suffers similarly to the latter in cases of slippery terrain since the vertical GRF might assume large values. 
Rotella \textit{et al.}~\cite{RotellaContact} proposed an unsupervised learning framework that employs F/T and IMU measurements to perform clustering using fuzzy c-means. 
The sensors were mounted on the feet of a humanoid robot and they estimated the probability of contact for each one of the six Degrees of Freedom (DoFs) of the end effector.
Likewise, in~\cite{haptic_explore} the authors approach the quality of contact prediction by using unsupervised learning. In order to estimate the contact state, they measured the static friction coefficient by using haptic exploration of the ground with the robot's feet. Unsupervised learning was also employed to estimate the gait-phase probability in~\cite{PiperakisGEM}.

Contemporary learning based approaches have been shown efficient in practice, but as mentioned previously they require ground truth labels during the training process.
Camurri \textit{et al.}~\cite{Camurri} used a one-dimensional logistic regression framework in order to specify dynamic GRF thresholds for each gait type. 
The learnt threshold was applied only on the vertical GRF while the other components were ignored under the assumption of sufficient friction to prevent slippage. 
Recently, Lin \textit{et al.}~\cite{lin2021legged} introduced a deep convolutional neural network that utilizes IMUs and joint encoders to classify individual contacts as stable/unstable ones.
Their approach was assessed on a Mini Cheetah robot and across various terrains where they reported an overall 97\% accuracy. 
Similarly, in our previous work~\cite{lcd}, we proposed a deep learning framework that employs F/T and IMU measurements to extract the contact state.
We have shown that the framework can generalize and make predictions on different robots without providing any new data and with only slightly inferior results.
This capability indicates that there are contact invariant features that facilitate the generalization across robots with different sizes and weights. 
The latter constitutes the main motivation for the current work towards a generic, platform and environment agnostic method.

Most of the aforementioned approaches from contemporary literature focus on classifying the contact as a binary state, with some notable exceptions such as~\cite{RotellaContact,haptic_explore}. In order to exploit maximum information regarding the state of a foot, the contact state should not be treated as a categorical variable (contact/no contact), but rather as a continuous one.
In many scenarios in uneven and/or slippery terrains, it is possible for a foot to be in touch with the ground regardless of the fact that the contact is unstable, giving rise to devastating consequences if this is not taken into account during locomotion.
\subsection*{Paper Contributions}
In this work, we treat the contact state as a six DoF continuous variable and estimate each individual contact probability which is then fused into the final estimation. Our main contributions can be summarized as follows:
\begin{itemize}
    \item To the best of our knowledge this is the first work that estimates the contact probability  by considering solely inertial measurements.
    \item The proposed stable contact detection module is robot agnostic. In other words, it can be employed on any legged robot that is equipped with mid-range IMUs mounted on its feet. By executing a few steps on a surface with sufficient friction (ensure stable contact), the module can trivially be fine tuned for the employed robotic platform. 
    \item To further facilitate and promote research in the sector, both the code and the datasets that the proposed method was tested upon are released as an open-source project at~\cite{PCE}. 
\end{itemize}
Overall, robust contact detection can be exploited in various fields such as locomotion control~\cite{Herzog2016}, base state estimation~\cite{PiperakisIROS19,RotellaStateEstimation}, Center of Mass (CoM) estimation~\cite{RotellaMomentum} and also improve SLAM on legged robots~\cite{Hourdakis2021}.

The rest of this paper is structured as follows. In Section II, the contact detection problem is formulated alongside with the employed measurement model. Section  III describes the mathematical modelling of the proposed approach. Finally, Section IV is reserved for experimental results and Section V concludes this article with a discussion and directions for future work.

\begin{figure}[b]
    \centering
    \includegraphics[width=.95\columnwidth]{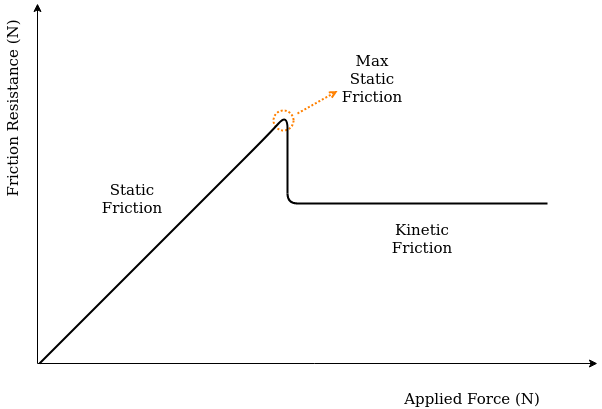}
    \caption{Static and Kinetic friction force.}
    \label{fig:statickinetic}
\end{figure}

\begin{figure*}[h]
    \centering
    \includegraphics[height= 5.5cm,width=.7\textwidth]{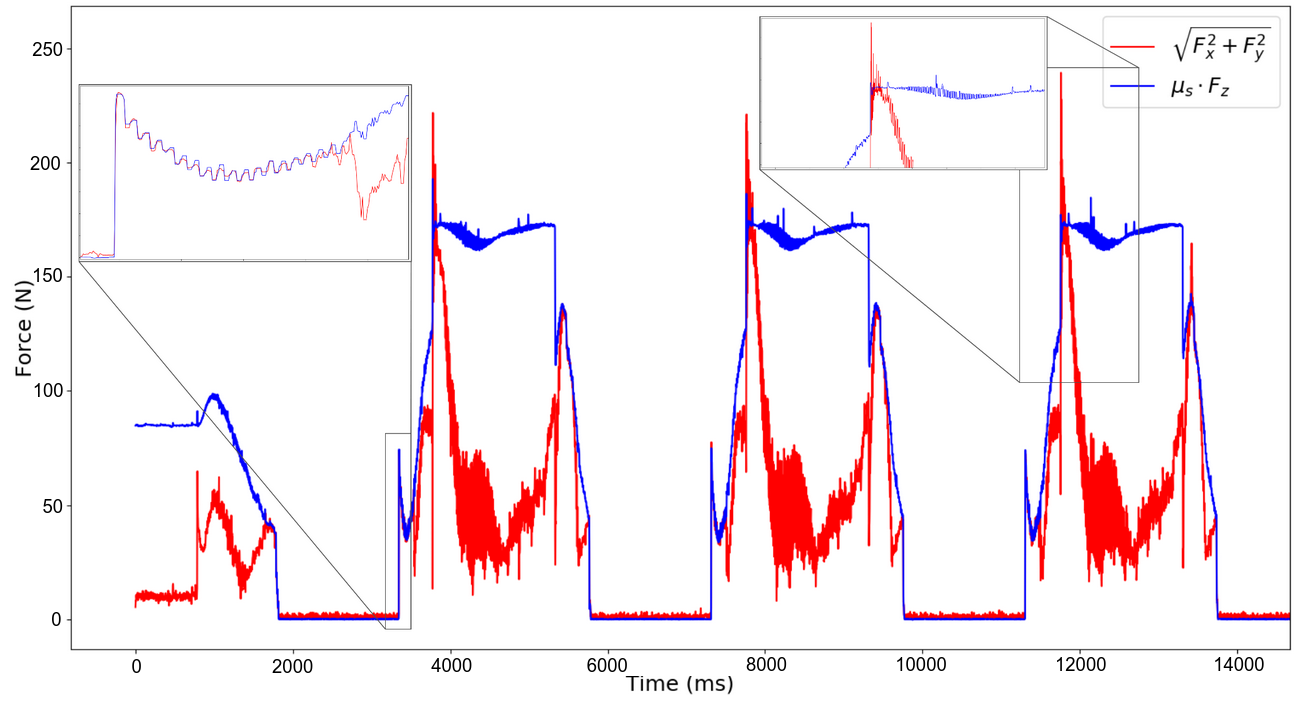}
    \caption{Force fluctuations during impact for the ATLAS simulated robot.}
    \label{fig:Force}
\end{figure*}

\section{Problem Formulation}
\subsection{Background}
Static friction is the force that prevents an object from moving when the relative speed between the object and the supporting surface is zero. Frictional force offers the resistance to the applied force opposing its motion:
\begin{equation}
    T_s = \mu_s F_z 
\end{equation}
where $T_s$ is the static friction force, $\mu_s$ is the static friction coefficient and $F_z$ is the normal to the plane GRF.
On the contrary, when an object is moving and is in touch with a surface, the kinetic friction force $T_k$ is resisting it's motion:
\begin{equation}
    T_k = \mu_k F_z
\end{equation}
where $\mu_k$ is the kinetic friction coefficient. 
As a general rule, $\mu_k$ is smaller than $\mu_s$ and hence $T_k<T_s$, indicating that once the object starts moving it is harder to stop, because the resistance friction force is smaller.
Although most contemporary approaches assume that the friction force is constant, in practice this is not the case.
In Fig.~\ref{fig:statickinetic}, we depict how the resistance force varies when the applied force assumes larger values.
In legged robot locomotion, when the threshold of motion is surpassed, the resistance force instantly reduces and the foot begins to accelerate for a certain period of time until either the robot loses balance entirely or deceleration occurs and the foot comes to rest. 
Although it is possible for an object to be moving with constant velocity $(a_x = a_y = a_z = 0 \frac{m}{s^2})$ when the applied and resistance forces are equal, for the reasons stated previously it is infeasible to occur on the robot's foot during dynamic locomotion.
Consequently, we explore the cases when the foot is completely still or it accelerates after the threshold of motion is surpassed.

According to the Coulomb's model for dry friction, in order for a contact to be classified as stable, i.e. the relative speed between two objects in contact is zero, the following condition must hold true:

\begin{equation}\label{eq:CoulombFriction}
\sqrt{F_x^2 + F_y^2} \leq \mu_s F_z , \: \: F_z > 0
\end{equation}
where $F_x$ and $F_y$ are the lateral forces at each contact point. Although in physics Eq.~(\ref{eq:CoulombFriction}) describes the contact state deterministically (stable or unstable), a problem arises during impact when vibrations occur between the foot and the ground while simultaneously the robot transfers its weight towards the support leg. These fluctuations between the two terms of Eq.~(\ref{eq:CoulombFriction}), namely $\sqrt{F_x^2 + F_y^2}$ and $\mu_s F_z$, are illustrated in Fig.~\ref{fig:Force}, where if the blue line is greater than the red one, the contact is considered stable.
As can be observed, during impact the contact state oscillates between stable and unstable until the robot transfers its weight and the absolute difference between the two terms becomes significant.
Accordingly, in this work we aim at detecting this part of the gait phase as will be illustrated in the experimental results section.

\subsection{Measurement Model}
The following model was considered for the IMU measurements:
\begin{align}
     \boldsymbol{\alpha}^f &=  \boldsymbol{\bar{\alpha}}^f + {}^f \boldsymbol{R}_w \boldsymbol{g}  + \boldsymbol{b}_\alpha + \boldsymbol{w}_\alpha\\
     \boldsymbol{\omega}^f &=   \boldsymbol{\bar{\omega}}^f + \boldsymbol{b}_\omega + \boldsymbol{w}_\omega
\end{align}
where $\boldsymbol{\alpha}^f \; \epsilon \; {\rm I\!R}^3 $ and $\boldsymbol{\omega}^f \; \epsilon \; {\rm I\!R}^3$ are the linear acceleration and angular velocity measurement vectors for the corresponding foot as measured by the IMU in the local foot frame, respectively. 
For a humanoid, $f\; \epsilon \; \{L,R\}$ (left or right), while for a quadrupedal $f \; \epsilon \; \{RL,RR,FL,FR\}$ (rear left, rear right, front left and front right). 
$\boldsymbol{\bar{\alpha}}^f$ and $ \boldsymbol{\bar{\omega}}^f$ are the respective true values, ${}^f \boldsymbol{R}_w$ is the rotation from world to the corresponding leg frame, $\boldsymbol{g}$ is the gravity vector and finally $\boldsymbol{b}_\alpha$, $\boldsymbol{b}_\omega$, $\boldsymbol{w}_\alpha$, and $\boldsymbol{w}_\omega$ are the biases and the zero-mean normally distributed noises.
Additionally, biases and the gravity constant are removed from the accelerometer and gyroscope in real-time by estimating the rotation (${}^f \boldsymbol{R}_{w}$) with a complementary filter on the IMU measurements~\cite{filter}. This simplifies the measurement model to the following:
\begin{align}
     \boldsymbol{\alpha}^f &=  \boldsymbol{\bar{\alpha}}^f +  \boldsymbol{w}_\alpha \label{eq:accel_model}\\
     \boldsymbol{\omega}^f &=   \boldsymbol{\bar{\omega}}^f  + \boldsymbol{w}_\omega \label{eq:angul_model}
\end{align}

\subsection{Stable contact definition}
As stated previously, at the moment of impact between foot and the ground, and for a small period of time after that, the forces fluctuate and micro-movements of the foot occur. The same issue arises during the transition from support leg to swing. 
A contact is defined as tangentially stable when the following conditions are satisfied:
\begin{align}
    \bar{v}_x^f &= 0 \label{eq:velx}\\
    \bar{v}_y^f &= 0 \label{eq:vely}\\
    \bar{\omega}_z^f &=0
\end{align}
where $\bar{v}_x^f$,$\bar{v}_y^f$ are the true velocities of the $f$ foot in the $x$ and $y$ axes, respectively. 
In the same manner, for the rotational stable state:
\begin{align}
    \bar{v}_z^f &= 0 \\
    \bar{w}_x^f &= 0 \\
    \bar{\omega}_y^f \label{eq:wy}&=0
\end{align}
As already explained above (subsection B), constant velocity is infeasible while slipping. Accordingly, Eqs.~(\ref{eq:velx})--(\ref{eq:wy}) can be reformulated to fit the IMU measurements as:
\begin{align}
    \bar{a}_x^f = \bar{a}_y^f = \bar{a}_z^f &= 0 \label{eq:ax}\\
    \bar{\omega}_x^f = \bar{\omega}_y^f = \bar{\omega}_z^f &= 0 \label{eq:ws}
\end{align}

\section{Mathematical Modeling}
Assuming an ideal scenario with no uncertainties in the measurements, Eqs.~(\ref{eq:ax}--\ref{eq:ws}) are sufficient to classify the contact state between the robot foot and the ground deterministically. However, the latter is far from being true in real cases. 
In this work, we exploit the probabilistic nature of involved uncertainties to extract the stable contact probability at each time step. For the relevant formulation we start with Eqs.~(\ref{eq:accel_model}) and (\ref{eq:angul_model}) which can be re-written as:\\
\begin{align}
    \boldsymbol{\bar{\alpha}}^f  &= \boldsymbol{\alpha}^f - \boldsymbol{w}_\alpha \\
     \boldsymbol{\bar{\omega}}^f &= \boldsymbol{\omega}^f - \boldsymbol{w}_\omega 
\end{align}
Since the uncertainty $\boldsymbol{w}$ is modelled as a zero-mean normally distributed random variable (r.v.), $\boldsymbol{w}\sim N(\boldsymbol{0},\sigma^2)$, the equations above describe the ground truth measurements as normally distributed r.v. too. 
In this context, $\boldsymbol{\bar{\alpha}}^f \sim N(\boldsymbol{\alpha^f},\sigma^2)$ and $\boldsymbol{\bar{\omega}}^f \sim N(\boldsymbol{\omega^f},\sigma^2)$, where $\boldsymbol{\alpha^f}$ and $\boldsymbol{\omega^f}$ stand for the probabilistic measurements from the IMU. 
Accordingly, the goal of the proposed approach can be defined as to find the probability of the ground values to be approximately zero for all six axes.  

To this end, we employ a Kernel Density Estimator (KDE) to approximate the Probability Density Function (PDF) that describes a batch of samples. In this work, the batch size has been experimentally set to 50. A good rule of thumb is to set the batch size to be an order of magnitude less that the sensor's refresh rate.

\begin{figure}[t!]
    \centering
    \includegraphics[width=.95\columnwidth]{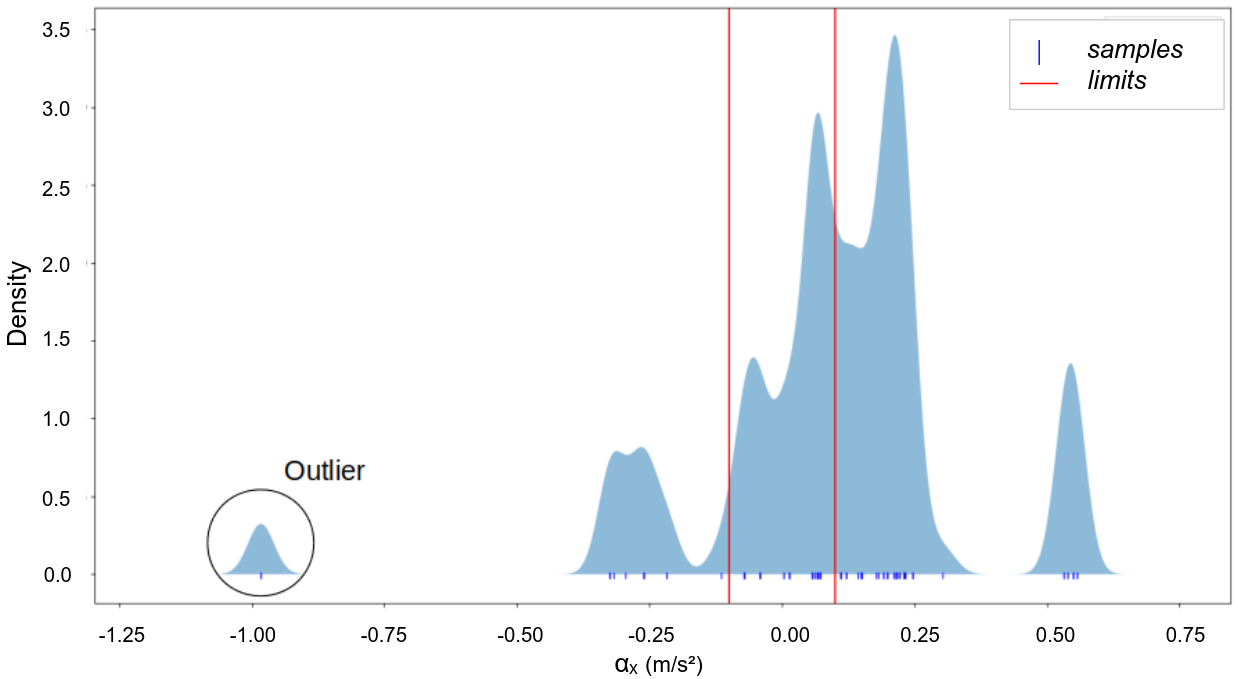}
    \caption{Example of estimated PDF for 100 $a_x$ samples.}
    \label{fig:kdewithoutlier}
\end{figure}

By integrating the estimated PDF over a symmetrical interval, the stable contact probability for each axis is calculated. Finally, since the per axis probabilities are independent, the final estimate is obtained  by multiplying the individual probabilities:
\begin{equation}
    P(stable|\boldsymbol{m}^t,\boldsymbol{m}^{t-1},..., \boldsymbol{m}^{t-d}) = \prod_{i =1}^{6} P(|\boldsymbol{m}_i^t| < \boldsymbol{\delta}_i)
\end{equation}
where $\boldsymbol{m} \; \epsilon \; {\rm I\!R}^6$ is the measurement vector, $d$ is the batch size and $\boldsymbol{\delta}\; \epsilon \; {\rm I\!R}^6$  is the empirical range vector for each axis of the IMU (red lines in Fig~\ref{fig:kdewithoutlier}) that indicates the interval at which the estimated PDF will be integrated.
When a new measurement is acquired, we use it along with the previous ones to estimate the updated PDF. 
KDE is a non-parametric estimator of univariate or multivariate densities with well-defined properties~\cite{kde}.
The reason we employed KDE over Maximum Likelihood Estimation (MLE) to approach the PDF is to take into consideration the transition from support leg to swing, which is only possible if the PDF is multimodal. 

By employing the Markov assumption and the measurement model that is presented in Section II we can formulate KDE in order to fit our case. 
Let $m_1, m_2, ... m_n$ be independent and identically distributed IMU measurements, $m \; \epsilon \; {\rm I\!R}^6 $. The density function is described as follows:
\begin{equation}
    \hat{f}_h(x) = \frac{1}{nh}  \sum_{i=1}^{n} K\left(\frac{m-m_i}{h}\right)
\end{equation}
where $\hat{f}$ is the PDF, $n$ is the number of samples, $h$ is termed bandwidth and $K$ is the Gaussian kernel:
\begin{equation}
    K(u) = \frac{1}{\sqrt{2\pi}}exp\left(-\frac{u^2}{2}\right)
\end{equation}
The bandwidth parameter $h$ describes how wide the PDF of every individual normally distributed sample will be represented. Naturally, we've chosen this parameter to be equal to the standard deviation that each sensor's specification sheet provides. After the PDF is estimated we integrate over a small interval ($[-\boldsymbol{\delta},\boldsymbol{\delta}]$) to compute the probability.

In Fig.~\ref{fig:kdewithoutlier}, we present the 1-axis PDF that was estimated over a batch of 100 $a_x$ samples of the ATLAS humanoid robot in simulation. These samples are carefully selected to depict the beginning of the foot's transition from support to swing. In any instance, that the foot is completely still and all measurements are gathered around zero, the PDF is roughly a normal distribution with mean $\mu \thickapprox 0$. It is worth stating that KDE is outlier robust since few measurements do not greatly affect the overall distribution.

\section{Experimental Results}
The current section is dedicated to the experimental evaluation of the proposed method. Detailed testing and evaluation has been conducted by employing three different robotic platforms as well as various terrains and friction conditions.
More specifically, two humanoids and one quadruped robot were used, demonstrating the generality and broad applicability of our method. 
For each robotic platform a number of experiments were performed which are reported and documented in the following. In addition, we have exploited the simulation environment in order to quantitatively and comparatively assess our method. In addition to the above,
a more detailed illustration of our experiments is presented in high resolution at \url{https://youtu.be/2CEkifEAQEc}. 
Throughout this section, we explore the contact state of one foot. 
Since each foot of the robot is identically constructed, the same process can be readily applied to the rest of the feet.

\subsection{Simulation Results}
The first set of experiments was conducted with a simulated ATLAS humanoid robot in a highly accurate multi-contact simulation environment namely RaiSim~\cite{raisim}. We generated omni-directional walking patterns by employing a stabilization module~\cite{LIPMControl,ScaronControl,wholebodyIK} based on Linear Inverted Pendulum (LIP) dynamics.
The refresh rate of the IMU that is mounted on the sole of the robot is $1000Hz$. The standard deviations of the zero-mean Gaussian noises are $\sigma_a = 0.02467 \frac{m}{s^2}$ and $\sigma_\omega =0.01653 \frac{rad}{s} $ for linear acceleration and angular velocity, respectively.

\begin{figure}[h]
    \centering
    \includegraphics[width=.95\columnwidth]{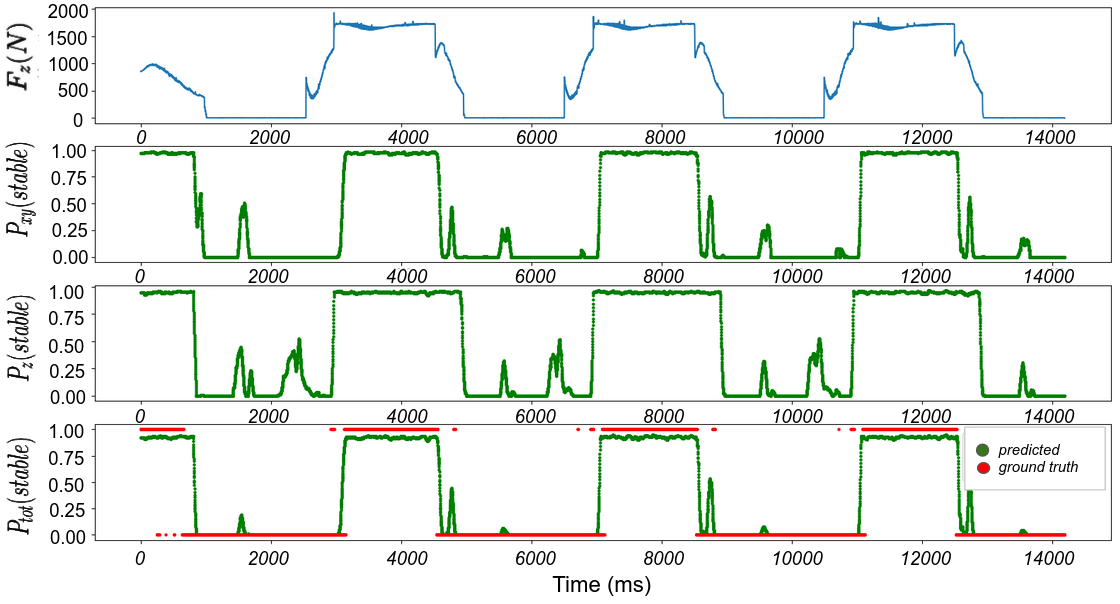}
    \caption{ATLAS walking on stable surface.}
    \label{fig:atlassimprob}
\end{figure}

In a first experiment, the friction coefficient was assumed constant ($\mu_s = 0.1$) and we explored the following gait pattern: Double Support (DS) and three consecutive right footsteps, i.e. Right Single Support (RSS).
In Fig.~\ref{fig:atlassimprob} we illustrate the vertical GRF as a point of reference, the tangential stable contact probability (parallel to the walking plane), the vertical probability and finally in the same sub-figure the total predicted probability with the ground truth labels.
The ground truth labels (stable/unstable) are extracted by taking into account the velocity of the sole and the vertical GRF. 
When the vertical GRF is greater than zero and the velocity norm of the sole is zero then the contact is classified as stable.

\begin{figure}[h]
    \centering
    \includegraphics[width=.95\columnwidth]{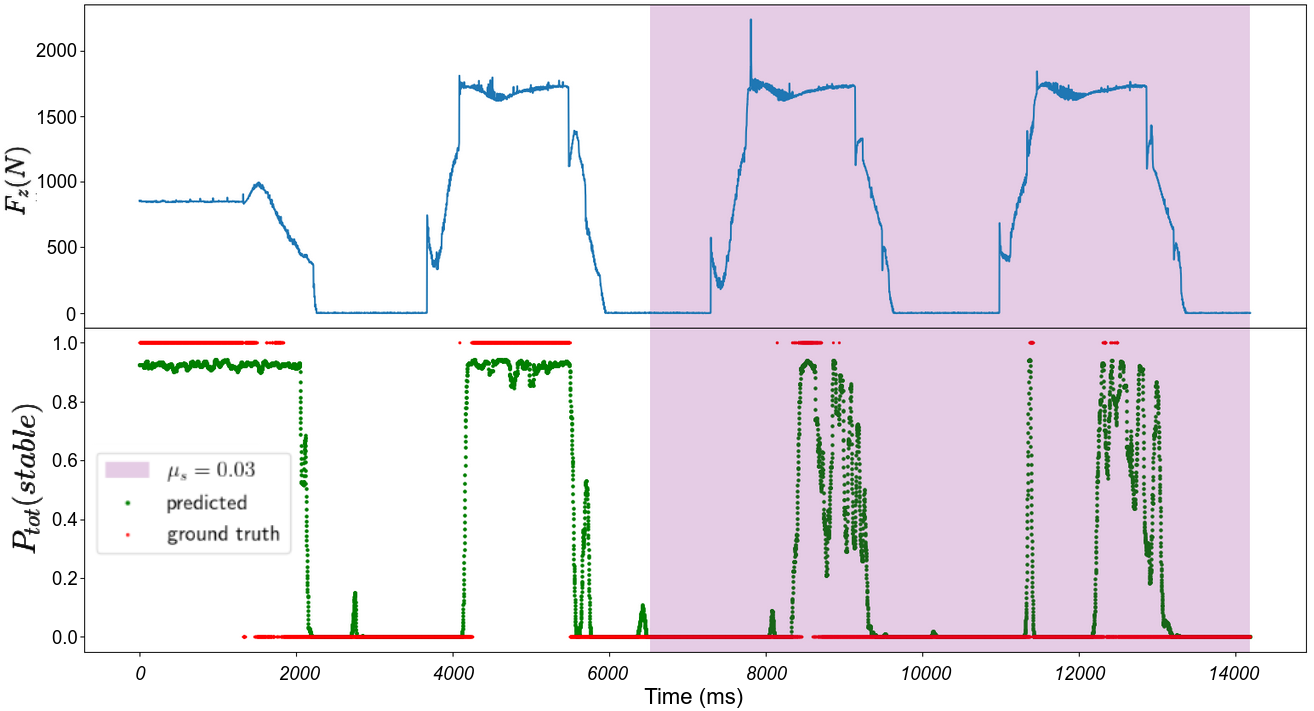}
    \caption{Estimated stable contact probabilities for ATLAS gaits on slippery terrain.}
    \label{fig:atlasslip}
\end{figure}

Next, we added a slippery surface with $\mu_s = 0.03$ and we commanded the robot to walk over it.
The robot starts at DS, then the first RSS is on a stable surface with $\mu_s = 0.1$ while the last two are on a slippery one ($\mu_s = 0.03$). In Fig.~\ref{fig:atlasslip}, purple region, we demonstrate the behavior of the stable contact probability when the foot is slipping.
Our predictions are confirmed by the ground truth labels. Please note that by simply thresholding the vertical GRF, the purple region steps would be misclassified as they are identical to the first one.

\subsection{Real experiments}
\begin{figure}[h]
    \centering
    \includegraphics[width=.95\columnwidth]{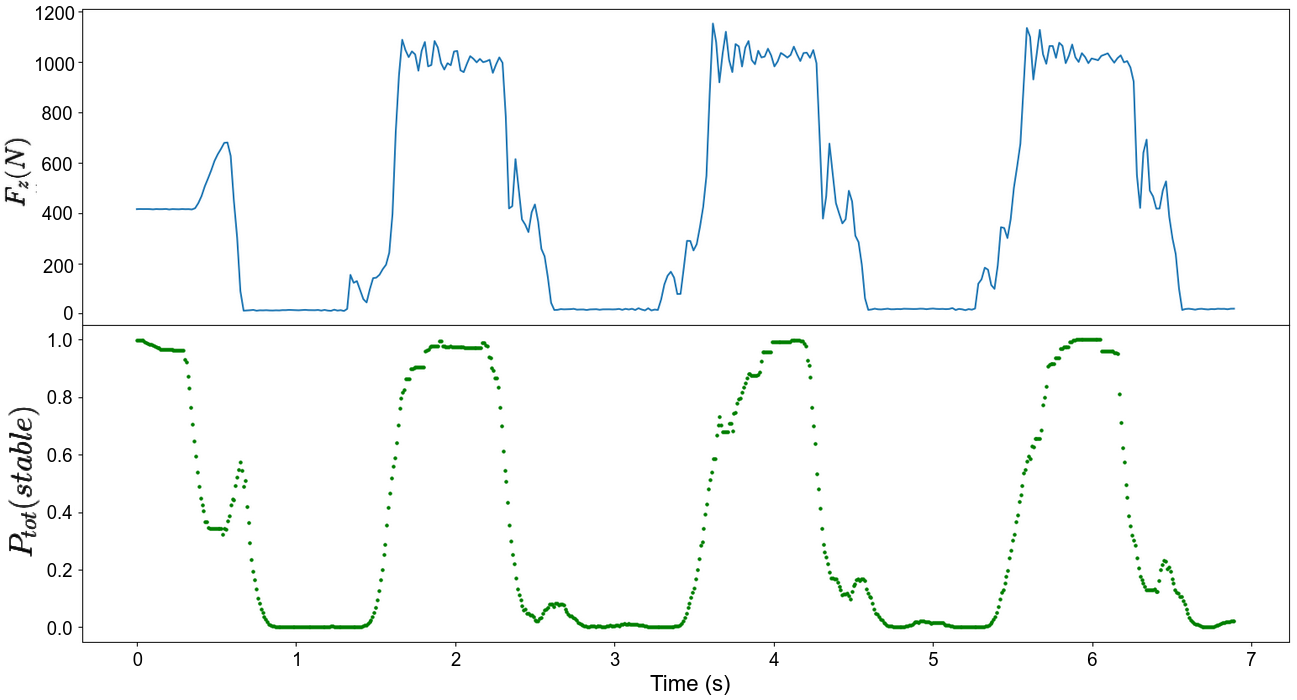}
    \caption{TALOS walking on a stable surface.}
    \label{fig:talosreal}
\end{figure}

The current section presents experimental evaluation of the proposed method with two robotic platforms, namely a TALOS humanoid and a GO1 quadruped robot. In a first experiment, the TALOS robot walked a few steps on a flat surface with sufficient friction coefficient to prevent slippage. 
The gait pattern is almost indistinguishable to the simulated ATLAS, since both robots are full sized humanoids with LIP-based walking pattern generation. 
The obtained results are presented in Fig.~\ref{fig:talosreal}. As can be observed, the contact estimation probabilities follow the expected pattern which also happens to be very similar to the one observed in the case of the ATLAS robot (Fig.~\ref{fig:atlassimprob}).

\begin{figure}[h]
    \centering
    \includegraphics[width=.95\columnwidth]{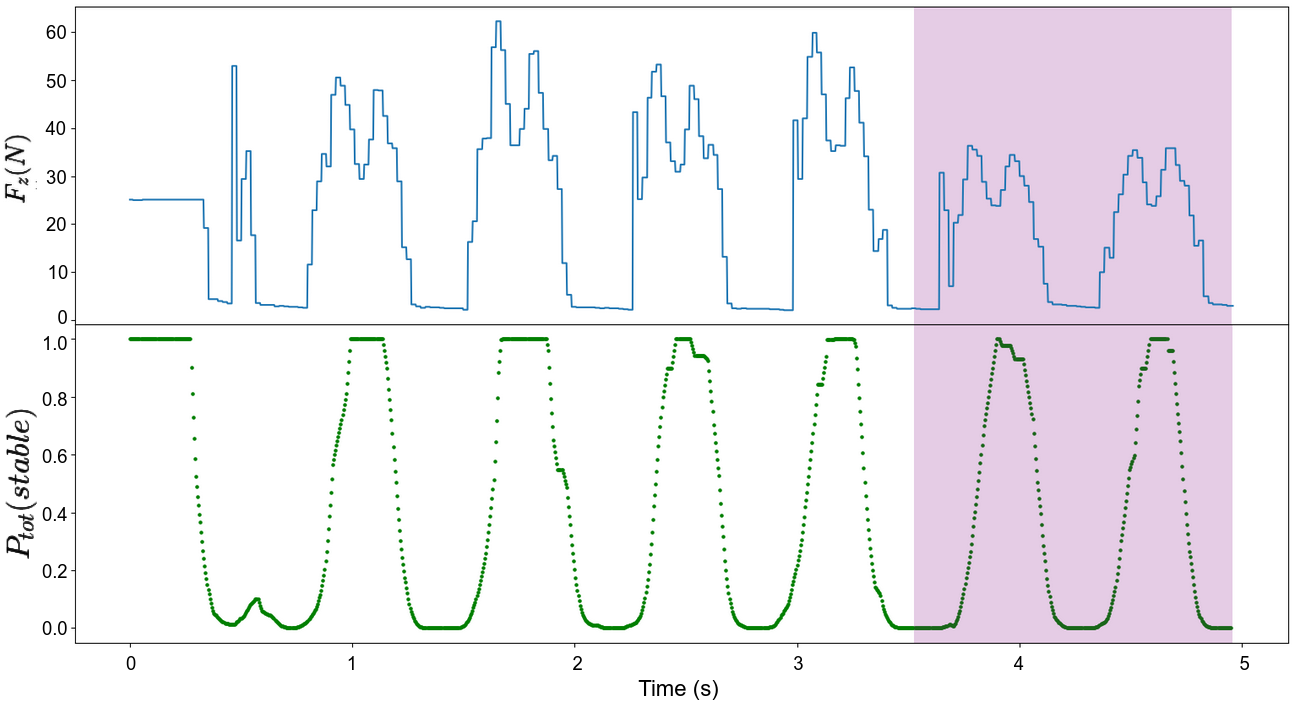}
    \caption{GO1 walking on soft terrain/low restitution.}
    \label{fig:go1matress}
\end{figure}

Subsequently, we evaluated extensively our method with the quadrupedal GO1 in various scenarios such as soft terrain and extremely slippery surface. 
In the first experiment, we commanded the robot to walk over a mattress with the default controller as provided by Unitree, to test the behavior of the contact probability when the vertical GRF decreases due to low restitution. 
The GRF measurement was acquired from a pressure sensor mounted on the robot's end effector and was used just as a point of reference. 
The inertial measurements that were used for estimating the contact probability were acquired by a low-end IMU sensor (LSM6DSOX) that was manually mounted on the GO1's end effector.
The purple region in Fig.~\ref{fig:go1matress} signifies the steps on the soft terrain.
As verified by the lower plot, the contact probability estimates remain unaffected by the terrain change. 
On the contrary, it is worth noting that approaches that utilize GRF measurements to classify the contact state would be greatly affected by the decrease of $F_z$.

\begin{figure}[h]
    \centering
    \includegraphics[width=.95\columnwidth]{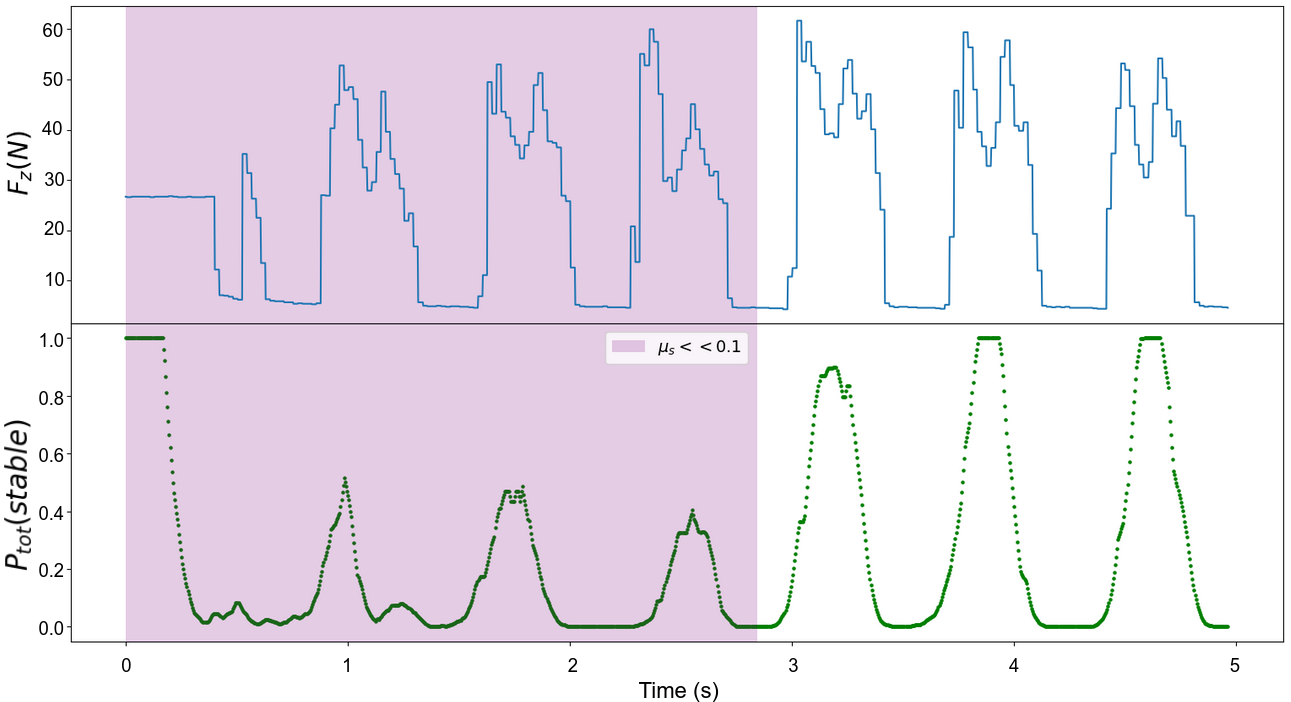}
    \caption{GO1 walking on slippery terrain for 3 consecutive  steps.}
    \label{fig:go1slip}
\end{figure}

For our final experiment, we greased the smooth surface that the robot was walking on to provoke extreme foot slippage. The purple region of Fig.~\ref{fig:go1slip} contains the extreme unstable steps of the robot before the grease wears out and the contacts are stable again. 
As Fig.~\ref{fig:go1slip} clearly shows, the contact probabilities in the greasy area are way lower indicating slippery behavior.

The proposed method can be employed in real-time with over 500 Hz refresh rate (performance measured in our implementation with a mid-range PC). We used the vertical GRF measurements to distinguish the following cases.
First, when the stable probability prediction is small, the foot can be either in (a) swing phase ($F_z = 0$) or (b) in unstable contact with the ground ($F_z >0$). Second, when the stable probability prediction is substantial, the foot (a) might be experiencing very small (close to zero) linear acceleration and angular velocity during swing ($F_z = 0$) or (b) is in stable contact state ($F_z > 0$). 
Nevertheless, since the vertical GRF is only considered to determine if the foot is in contact with the ground or not, an alternative would be to employ pressure or haptic sensors, even gait planning information e.g. when the leg is planned to be in swing or stance.

\subsection{Comparative evaluation}
\begin{figure}[h]
    \centering
    \includegraphics[width=.95\columnwidth]{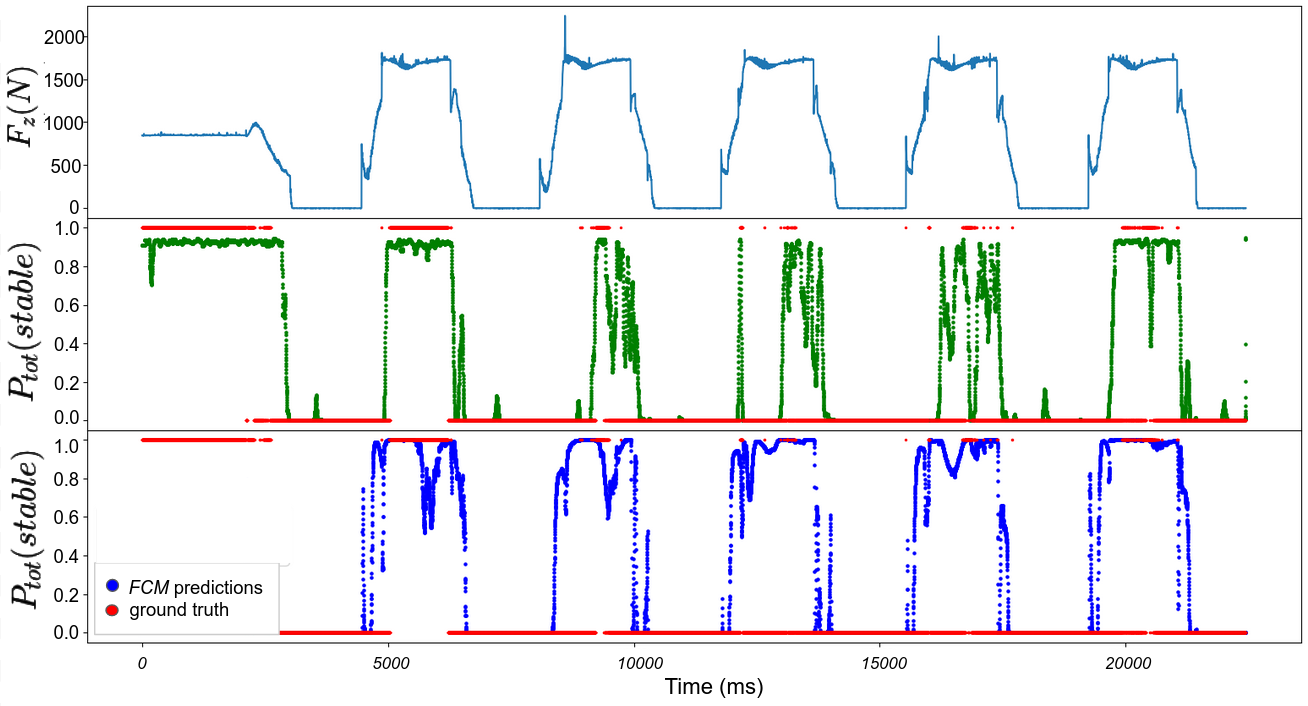}
    \caption{Comparative results; top row: vertical GRF; middle row: Contact probabilities as computed by proposed method; bottom row: Contact probabilities as computed by~\cite{RotellaContact}.}
    \label{fig:fcm}
\end{figure}

In order to quantitatively assess our method we have conducted comparative evaluation against a state of the art approach that probabilistically addresses the contact estimation problem~\cite{RotellaContact}. 
Fig.~\ref{fig:fcm} illustrates the contact probability estimations for the ATLAS simulated robot over a gaiting session that involved 22,000 discrete data points. 
The employed dataset is an extended version of the one previously used in Section~IV-A.
Overall, the experiment involved five footsteps with the middle three ones exhibiting highly unstable contact. Evidently, our method succeeded in correctly estimating the relevant probabilities, whereas the \cite{RotellaContact} approach only partially accomplished the task. The latter can be interpreted by the fact that~\cite{RotellaContact} depends on the full F/T measurements which are misleading. The results shown in Fig~\ref{fig:fcm} can be compactly represented by their RMSE values, indicating superior performance for our method: 
\begin{itemize}
    \item FCM approach~\cite{RotellaContact}: RMSE = 0.6076
    \item Proposed method:\hspace{0.35cm} RMSE = 0.3529
\end{itemize}

\begin{figure}[h]
    \centering
    \includegraphics[width=.97\columnwidth, height=3.3cm]{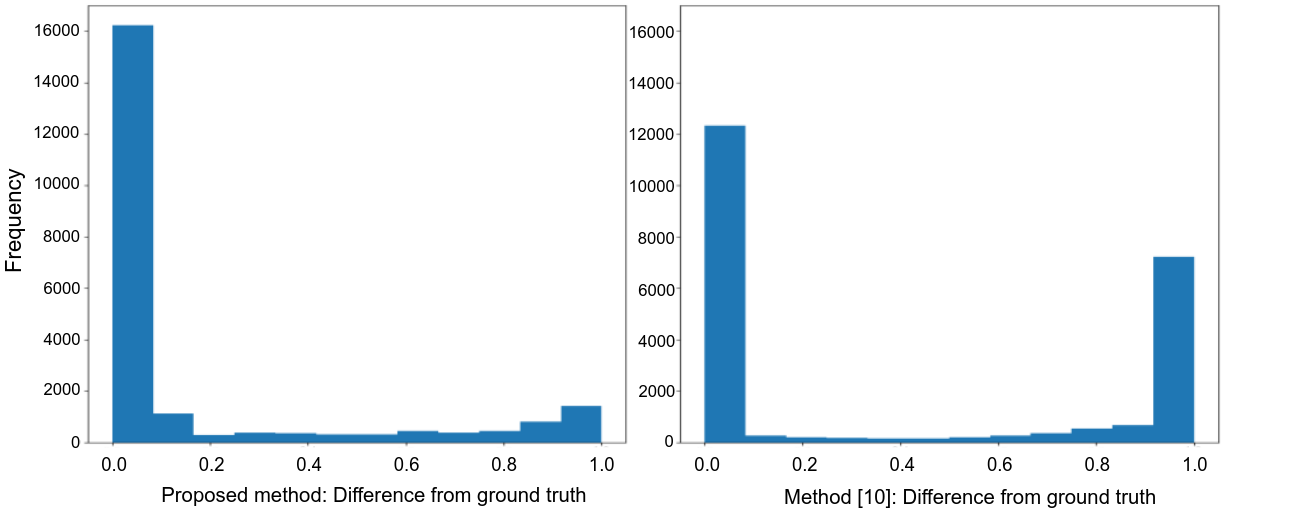}
    \caption{Histograms of the difference between predicted contact probability and ground truth labels. }
    \label{fig:hist}
\end{figure}
To better appreciate the quality of the estimated contact probabilities for both approaches, Fig.~\ref{fig:hist} illustrates the obtained histograms of the absolute differences between ground truth and predicted probabilities. 
As can be observed, the number of correct predictions is significantly larger for the proposed method whereas more erroneous predictions are exhibited in~\cite{RotellaContact}.

\section{Conclusions}
We have demonstrated that the proposed probabilistic contact estimator predicts successfully the contact state of legged robots, both in simulation and in real platforms.
By employing only inertial measurements, we have shown that it can generalize in different scenarios, even in cases with extremely low static friction coefficient and successfully estimate the quality of contact.
Our presentation has revealed a number of important features that are inherent in our method:
\begin{itemize}
    \item Proprioceptive sensing is the sole information source that is employed; specifically, measurements from low-cost IMUs constitute the sensory input.
    \item No training data and ground truth labels are required, hence facilitating the operation of the method in practically any scenario.
    \item Equally important, the method is robot agnostic, making it a compelling contact estimation module for legged robots.
\end{itemize}
Overall, this approach provides a unified solution to the quality of contact detection problem, being at the same time robot and environment agnostic. 
In future work, we aim at conducting detailed sessions outside the lab environment, including outdoor terrains. In addition, we aim at integrating the proposed method with previously developed modules, such as legged robot state estimation and gait generation.

\section*{Acknowledgement}
The authors would like to thank Olivier Stasse, LAAS-CNRS, for providing the dataset for the TALOS experiment.

\bibliographystyle{IEEEtran}
\bibliography{root}

\end{document}